\documentclass[conference]{IEEEtran}
\IEEEoverridecommandlockouts
\usepackage{cite}
\usepackage{amsmath,amssymb,amsfonts}
\usepackage{algorithmic}
\usepackage{graphicx}
\usepackage{textcomp}
\usepackage[table,xcdraw]{xcolor}
\usepackage{comment}
\usepackage[ruled,vlined,linesnumbered]{algorithm2e}
\usepackage{multirow}
\usepackage{hhline}
\usepackage{caption}
\usepackage[caption=false]{subfig}
\usepackage{parskip} 

\usepackage[strings]{underscore}

\makeatletter
\newcommand{\removelatexerror}{\let\@latex@error\@gobble}
\makeatother
\makeatletter

\def\BibTeX{{\rm B\kern-.05em{\sc i\kern-.025em b}\kern-.08em
    T\kern-.1667em\lower.7ex\hbox{E}\kern-.125emX}}
\begin{document}

\title{Population Age Group Sensitivity for COVID-19 Infections with Deep Learning}

\makeatletter 
\newcommand{\linebreakand}{%
  \end{@IEEEauthorhalign}
  \hfill\mbox{}\par
  \mbox{}\hfill\begin{@IEEEauthorhalign}
}
\makeatother 

\author{}
 \author{\IEEEauthorblockN{Md Khairul Islam}
 \IEEEauthorblockA{\textit{Computer Science Department} \\
 \textit{University of Virginia} \\
 Charlottesville, USA \\
 mi3se@virginia.edu}
\and
 \IEEEauthorblockN{Tyler Valentine}
 \IEEEauthorblockA{\textit{School of Data Science} \\
 \textit{University of Virginia}\\
 Charlottesville, USA \\
 xje4cy@virginia.edu}
\and
 \IEEEauthorblockN{Royal Wang}
 \IEEEauthorblockA{\textit{School of Data Science} \\
 \textit{University of Virginia}\\
 Charlottesville, USA \\
 rjw8ng@virginia.edu}
 
\linebreakand

 \IEEEauthorblockN{Levi Davis}
 \IEEEauthorblockA{\textit{School of Data Science} \\
 \textit{University of Virginia}\\
 Charlottesville, USA \\
 ljd3frf@virginia.edu}
\and
 \IEEEauthorblockN{Matt Manner}
 \IEEEauthorblockA{\textit{School of Data Science} \\
 \textit{University of Virginia}\\
 Charlottesville, USA \\
 xkv3na@virginia.edu}
\and
 \IEEEauthorblockN{Judy Fox}
 \IEEEauthorblockA{\textit{Computer Science Department} \\
 \textit{School of Data Science} \\
 \textit{University of Virginia}\\
 Charlottesville, USA \\
 cwk9mp@virginia.edu}
 }

\maketitle
\begin{abstract}
The COVID-19 pandemic has created unprecedented challenges for governments and healthcare systems worldwide, highlighting the critical importance of understanding the factors that contribute to virus transmission. This study aimed to identify the most influential age groups in COVID-19 infection rates at the US county level using the Modified Morris Method and deep learning for time series. Our approach involved training the state-of-the-art time-series model Temporal Fusion Transformer on different age groups as a static feature and the population vaccination status as the dynamic feature. We analyzed the impact of those age groups on COVID-19 infection rates by perturbing individual input features and ranked them based on their Morris sensitivity scores, which quantify their contribution to COVID-19 transmission rates. The findings are verified using ground truth data from the CDC and US Census, which provide the true infection rates for each age group. The results suggest that young adults were the most influential age group in COVID-19 transmission at the county level between March 1, 2020, and November 27, 2021. Using these results can inform public health policies and interventions, such as targeted vaccination strategies, to better control the spread of the virus. Our approach demonstrates the utility of feature sensitivity analysis in identifying critical factors contributing to COVID-19 transmission and can be applied in other public health domains.

\end{abstract}

\begin{IEEEkeywords}
Sensitivity Analysis, Morris Method, Deep Learning for Time Series, Temporal Fusion Transformer, County-level COVID-19
\end{IEEEkeywords}

\section{Introduction}
The COVID-19 pandemic has posed significant challenges for governments and healthcare systems worldwide, highlighting the need for effective measures to manage and control the virus's spread. Understanding the factors that contribute to disease transmission is crucial for developing targeted public health policies and interventions. Age is a critical factor in COVID-19 transmission as shown by previous studies \cite{bae2021impact} \cite{dessie2021mortality}. 

Many prior works have studied forecasting COVID-19 infection and mortality rates using different statistical \cite{yang2020modified}, auto-regressive machine learning \cite{arunkumar2021forecasting}\cite{kirbacs2020comparative}, and deep learning \cite{zeroual2020deep}\cite{ramchandani2020deepcovidnet}\cite{arik2022self} models. The focus of these models is to make better predictions for the epidemic spread which would help take better mitigation steps. Interpreting how these models make these predictions can provide an understanding of the importance of these different input factors and their interactions \cite{ramchandani2020deepcovidnet}. Temporal Fusion Transformer (TFT) \cite{lim2021temporal} is a state-of-the-art forecasting model that has been widely used to do multivariate multi-horizon forecasting. TFT is particularly useful for modeling complex, non-linear relationships between input features and target variables, making it an ideal tool for analyzing COVID-19 transmission patterns.

The Morris method \cite{morris1991factorial} has been widely used to perform sensitivity analysis of different models \cite{aumond2021global}\cite{tsvetkova2021review}. By calculating the output change with reference to input perturbation, the Morris method provides a simple approach to rank input factors by their sensitivity. By combining TFT with feature sensitivity analysis, we can gain a better understanding of the factors that contribute to COVID-19 transmission and develop more effective strategies to control the spread of the virus. 

In this work, we collected the population by age subgroups data (Fig. \ref{fig:weekly_ground_truth}) for each of the 3,142 US counties\cite{pop2020}, along  with the daily vaccination rate of the population \cite{vaccination} and COVID-19 case report from March 1, 2020, to Dec 27, 2021 \cite{usafacts}. While there are prior studies that observe COVID-19 trends for different age groups \cite{monod2021age}, we trained a TFT model on the dataset separately on each age subgroup as the static covariate and then calculated the sensitivity of those models using a modified Morris method \cite{morris1991factorial}. We then ranked the age groups based on their Morris sensitivity scores to identify the most influential factors contributing to COVID-19 transmission. Finally, we evaluated our age group sensitivity rank with the actual COVID-19 cases reported for those age groups in the same time period \cite{weekly_cases_by_age}. Our study aims to provide insights into the age-related factors that contribute to COVID-19 transmission and inform targeted interventions to control the spread of the virus. 

\begin{figure}[ht]
\includegraphics[width=0.48\textwidth]{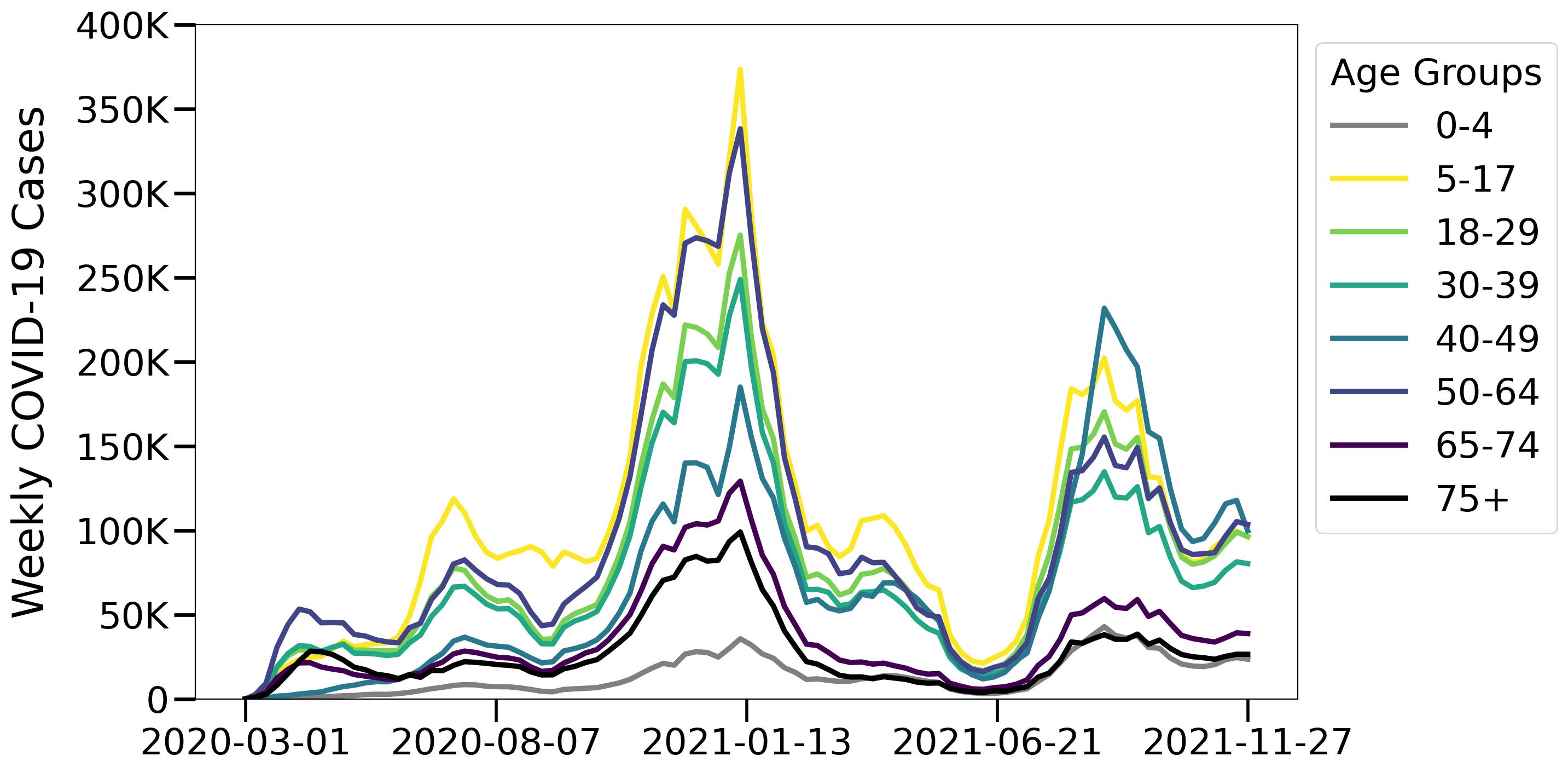}
\caption{Ground truth: weekly COVID-19 case numbers for each of the eight age subgroups from the time period of March 1, 2020, to November 27, 2021.}
\label{fig:weekly_ground_truth}
\end{figure}

The rest of the paper is organized as follows: Section \ref{sec:background} gives the background on the problem state and theoretical foundation of our work. Section \ref{sec:exp_setup} presents the experimental setup and model results. Section \ref{sec:sensitivity} features the age group ranking by Morris sensitivity score and comparison with ground truth. Section \ref{sec:related_works} summarizes the related works. Section \ref{sec:conclusion} contains the concluding remarks and potential future works.

\begin{figure*}[ht]
    \centering
    \includegraphics[width=0.80\textwidth, height=0.25\linewidth]{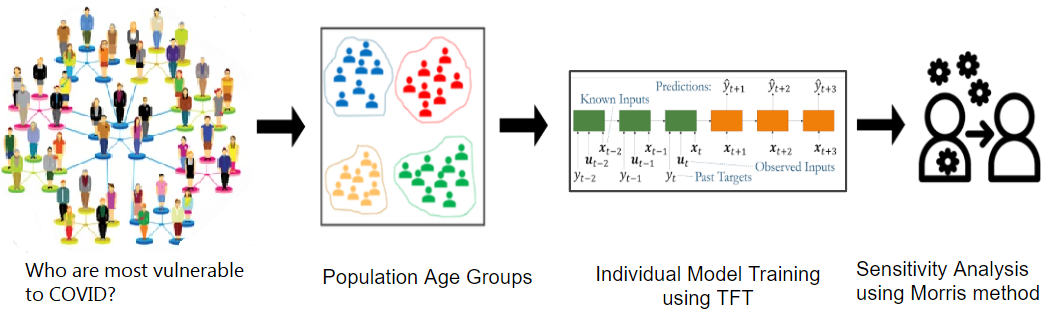}
    \caption{Workflow of Population Age Group Sensitivity Analysis for COVID-19 Forecasts.}
    \label{fig:Sensitivity_Analysis}
\end{figure*}

\section{Background}\label{sec:background}

\subsection{Sensitivity Analysis using the Morris Method}

Morris method is a reliable and efficient sensitivity analysis method that defines the sensitivity of a model input as the ratio of the change in an output variable to the change in an input feature. Given a model $f(.)$, and a set of $k$ input features $X (x_1, \dots, x_{k})$, the Morris sensitivity \cite{morris1991factorial} of a model input feature $x_{i}$ can be defined as follows:

\begin{equation}
\label{eqn:morris}
    Sensitivity(\boldsymbol{X}, i)=\frac{f(x_{1},\dots,x_{i}+\Delta,\dots,x_{k}) - f(\mathbf{X})}{\Delta}
\end{equation}

where $\Delta$ is the small change to the input feature $x_i$. Since the original Morris method was for static variables, we expand it for our predictions for the high dimensional spatial and temporal datasets. 

\textbf{Algorithm \ref{alg:norm_morris}} shows the implementation of the modified Morris method for our study, where we normalized the output value change by the number of input days, counties, and delta $\Delta$. Which we call the normalized Morris index $\hat{\mu*}$. We further scaled this index using the standard deviation ($\sigma$) of the input feature $x_{i}$, which we call the scaled Morris index ($\hat{\mu*} \times \sigma$). This scaling takes the underlying distribution of the feature $x_{i}$ when ranking the features by sensitivity. In the rest of the study, by Morris score, we refer to the scaled Morris index.


\SetKwComment{Comment}{/* }{ */}
\begin{figure}[!ht]
\small
    \renewcommand{\algorithmicrequire}{\textbf{Input:}}
    \renewcommand{\algorithmicensure}{\textbf{Output:}}
    \removelatexerror
    \begin{algorithm}[H]
        \label{alg:norm_morris}
        \caption{Modified Morris score calculation for \textbf{spatio-temporal} data}
        \KwIn{$\boldsymbol{X}=\{x_{1},\dots,x_{k}\}$, target feature $x_{i} \in \boldsymbol{X}$ with dimension $[C, T]$, model $f$, $\Delta$}
        \tcp*[h]{$\boldsymbol{X}$ is a set of $k$ input features, $\Delta$ is the change to $x_{i}$} \\        
        {$\boldsymbol{Y} = f(\boldsymbol{X})$} \\
        {$\boldsymbol{Y}_{\Delta} = f(x_{1},\dots,x_{i}+\Delta,\dots,x_{k})$} \\
        {$G, t = 0, 0 $} \\
        \While (/* Temporal */) {$t < T$} {
            \tcp*[h]{Loop through 637 Days} \\
            {$c = 0$} \\
            \While (/* Spatial */) {$c < C$} {
                \tcp*[h]{Loop through 3142 US Counties} \\
                {$G = G + \boldsymbol{Y}_{\Delta}[c][t] -  \boldsymbol{Y}[c][t]$} /*Total Change*/\\
                {$c = c+1 $} \\
            }
            {$t = t+1 $} \\
            {$c = 0 $} \\
        }
        \tcp*[h]{Calculate normalized Morris Index $\hat{\mu^{*}}$} \\
        {$\hat{\mu^{*}} = G / (C*T*\Delta)$}  \\
        \tcp*[h]{Get standard deviation of the feature} \\
        {$std = \text{standard_deviation} (x_i)$} \\
        \algorithmicreturn ~  $\hat{\mu^{*}} \times std$
    \end{algorithm}
    \label{fig:morris}
\end{figure}


\section{Problem Statement}
In this paper, we use deep learning to study feature sensitivity for model predictions of COVID-19 infection at the county level. The workflow for the design of our experiments is illustrated in Fig. \ref{fig:Sensitivity_Analysis}. 

\subsection{Model Training for Predictions using TFT \label{sec:tft}}

The Temporal Fusion Transformer (TFT) \cite{lim2021temporal} is a novel, interpretable, attention-based deep learning model for multi-horizon forecasting. Its architecture is carefully designed to handle static (e.g. percentage of \textit{age group} in a county population), past observed (e.g.  \textit{infection rate of cases}, \textit{daily vaccination coverage}), and known future (e.g. day of the week) inputs.  Its architecture is specially modified for four main uses. 1) To learn both local and global time-varying relationships at different scales. 2) To filter out input noises using Variable Selection Network (VSN). 3) To incorporate static metadata into the dynamic features for temporal forecasting. 4) To efficiently use different parts of the network through a gating mechanism. On a wide range of real-world tasks, TFT achieves state-of-the-art performance. Hence we use the model for our own study here. 

The prediction model $f(.)$ is denoted as bellow:

\begin{equation}
    \label{eqn:forecast}
    \begin{aligned}
    \hat{y}_i(t,\tau) = f(\tau, y_{i,t-k:t},\mathbf{z}_{i,t-k:t},\mathbf{x}_{i,t-k:t+\tau},\mathbf{s}_i)
    \end{aligned}
\end{equation} 

where $\hat{y}_i(t,\tau)$ is the  number of cases predicted at time $t \in [0, T_i]$ for county $i$. $\tau$ is the forecast horizon, 15 days in our case. $T_i$ is the length of the time series period, which for our case is the same for each county. We use the previous 13 days (lag window $k$) of data to forecast the next 15 days.



\section{Experimental Setup}
\label{sec:exp_setup}

\subsection{Computational Resources}
We implement our TFT model with PyTorch \cite{tft_pytorch}. Then we conducted a performance evaluation of the model training on HPC clusters including the GPU nodes in Table \ref{table:environment1}. Each training epoch takes an average of 50 minutes on a GPU node with at least 32GB of RAM. Each Morris runs with a trained model, and with additional feature analysis that takes around 35 minutes per epoch.

\begin{table}[!ht]
\centering
\caption{Runtime environment and hardware specification.}
\small
\begin{tabular}{|c|c|c|l|}
\hline
\textbf{Driver} & \textbf{CUDA} & \textbf{Processor} & \textbf{NVIDIA GPU} \\ \hline
 \multirow{4}{*}{470.82.01} & \multirow{4}{*}{11.4} & \multirow{4}{*}{Intel Xeon} & A100-SXM4-40GB \\ \cline{4-4}
 & & &Tesla P100-PCIE \\ \cline{4-4}
 && &Tesla V100-SXM2  \\\cline{4-4}
 & & &Tesla K80 \\ \hline
\end{tabular}
\label{table:environment1}
\end{table}

\subsection{Input Data and Features}
The data set consists of daily COVID-19 cases for over 3,142 US counties with eight static features and one dynamic feature. The static features signify the percentage of all people in one of the eight age subgroups (0-4, 5-17, 18-29, 30-39, 40-49, 50-64, 65-74, and 75 and older) for all counties. The dynamic feature indicates the percentage of people who are fully vaccinated for all counties. 

The data set was split into training, validation, and testing sets. The training set includes the dates March 1, 2020, through November 27, 2021, the validation set includes the dates November 28, 2021, through December 12, 2021, and the testing set includes the dates December 13, 2021, through December 27, 2021.

\subsection{Model Training and Prediction}\label{sec:prediction}
A separate TFT model was trained for each age group, for a total of eight models. Since all static variables are presented together as context vectors, the separation ensures the identification of a specific subgroup's contribution without the interference of other correlated static variables. In addition to the static age feature, all TFT models include a dynamic vaccination status feature. An additional variable SinWeekly is also included, which represents weekly trends for changes in the number of new cases. The target variable for each model was the daily number of cases for each county. The daily predictions for each county were determined for the training, validation, and testing time periods. These predictions were compared to the observed values to obtain RMSE values. 
\begin{table}[!ht]
    \centering
    \caption{RMSE values for all TFT models.}
    \begin{tabular}{|l|c|c|c|l|}
    \hline
    \textbf{Model} & \textbf{Train} & \textbf{Validation} & \textbf{Test} \\ \hline
        0-4 & 56.49 & 82.34 & 192.30 \\ \hline
        5-17 & 60.65 & 80.25 & 199.40 \\ \hline
        18-29 & 56.89 & 82.24 & 193.80 \\ \hline
        30-39 & 56.95 & 82.22 & 199.10 \\ \hline
        40-49 & 52.45 & 80.65 & 197.70 \\ \hline
        50-64 & 56.70 & 82.53 & 200.40 \\ \hline
        65-74 & 56.06 & 81.52 & 199.40 \\ \hline
        75+ & 60.72 & 83.80 & 204.40 \\ \hline
    \end{tabular}
    \label{table:rmse}
\end{table}

The RMSE values are shown in Table \ref{table:rmse} for each of the eight models. The "Model" column indicates the age subgroup used as the static variable for that model. As expected, all models performed well when predicting the test data set and have comparable loss values.

\subsection{Sensitivity Analysis}\label{sec:sensitivity}

Next, our Modified Morris Method was applied to each model. This involved perturbing the age feature in each model by a range of positive and negative delta values. The chosen values ranged from -0.01 to 0.01 with steps of 0.001, disregarding the delta value equal to zero. The Morris indices for each of the delta values are shown below (Fig. \ref{fig:morris_age_groups}). Positive and negative values of delta indicate an increase and decrease in the age feature, respectively. Investigating both an increase and decrease in these features allowed for an understanding of how the Morris indices change between different delta values. Our results show that the Morris values are approximately stable for each age group across a range of values. 

\begin{figure}[!ht]
\centerline{\includegraphics[scale=0.55]{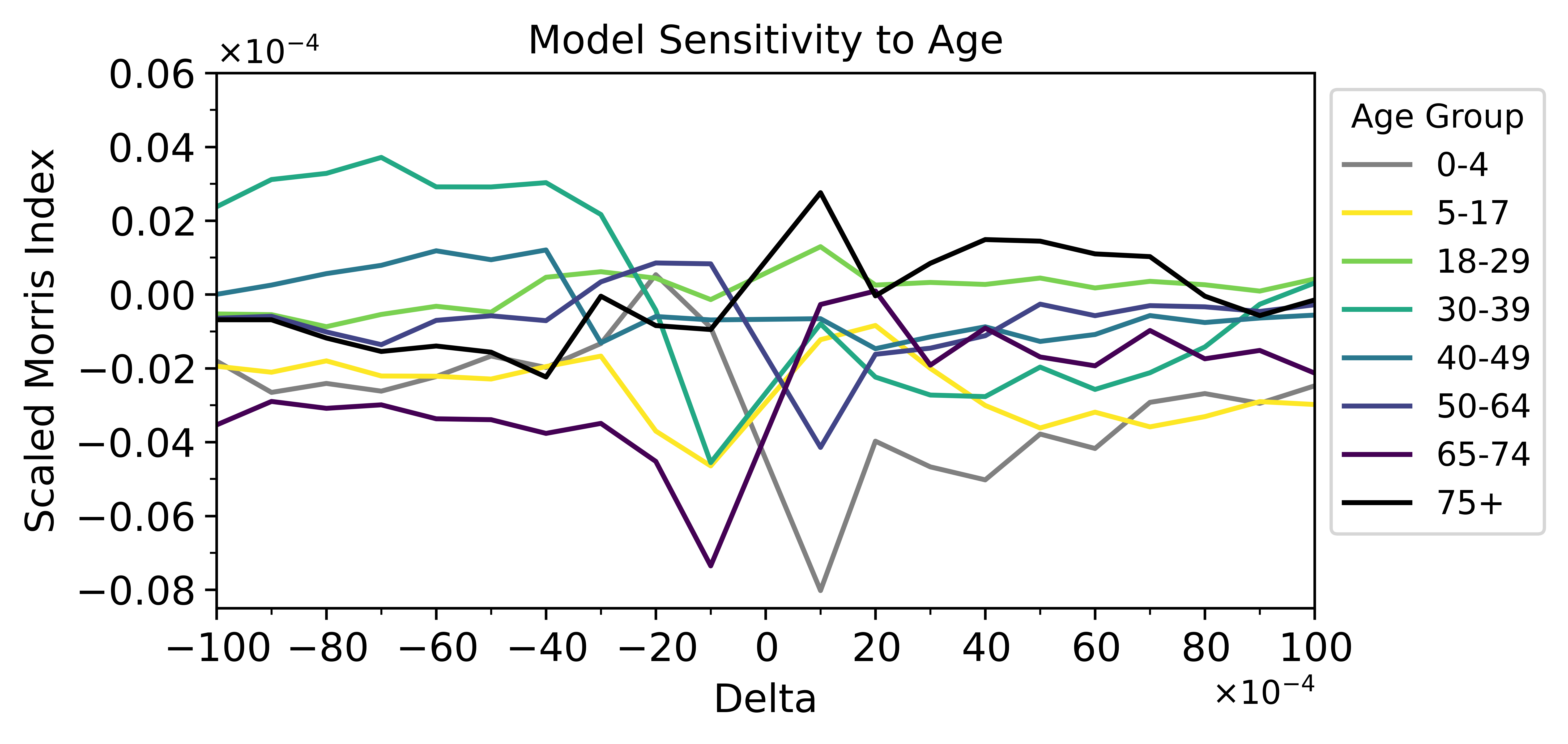}}
\caption{Morris indices as a function of delta for all eight models, calculated at intervals of 0.001, excluding zero.}
\label{fig:morris_age_groups}
\end{figure}

Another observation about these results is the magnitude of the Morris values. With the delta and Morris values shown at the same scale of $10^{-4}$, the Morris values remain about two to three orders of magnitude smaller than the delta values. This means that the changes caused by the perturbation of the static variable result in small changes in the predictions of the models, indicating a non-linear relationship between the static input and the output.

Despite the relatively small values, there is a clear distinction between which age groups have higher Morris indices on average. This separation allowed us to rank the age groups based on their sensitivity to the models. Before this was done, it was necessary to obtain a ground truth ranking to use for verification of our ranking derived from the Morris method. 

Data was obtained from the CDC for the total number of COVID-19 cases reported for each age group within the time period of the training set, as shown previously (Fig. \ref{fig:weekly_ground_truth}). Data from the US Census were also obtained to provide an estimate of the total population of each age group, with the estimate made for April 1, 2020. Having both the cases and the estimated population allowed us to calculate the true ranking. This was done by calculating the percentage of people in each age group who were infected. For example, 10,018,923 cases were reported for the age group 18-29, with an estimated population of 54,992,661 for that age group. This results in an infection rate of 18.2 percent. The infection rates for all age groups are shown below (Fig. \ref{fig:ground_truth}).


\begin{table}[!ht]
\caption{Total number of cases, total population, and infection rate of each age subgroup according to CDC and US Census data.}
\begin{tabular}{|c|c|c|c|c|}
\hline
\textbf{Age Group} & \textbf{Cases} & \textbf{Population} & \textbf{Infection Rate (\%)}                        \\ \hline
0 - 4 Years        & 1,249,223      & 19,392,551          & \cellcolor[HTML]{F7FBFF}{\color[HTML]{000000} 6.4}  \\
5 - 17 Years       & 6,184,296      & 54,992,661          & \cellcolor[HTML]{9AC8E0}{\color[HTML]{000000} 11.2} \\
18 - 29 Years      & 10,018,923     & 53,013,409          & \cellcolor[HTML]{08306B}{\color[HTML]{F1F1F1} 18.9} \\
30 - 39 Years      & 7,760,789      & 45,034,182          & \cellcolor[HTML]{0A539E}{\color[HTML]{F1F1F1} 17.2} \\
40 - 49 Years      & 6,767,348      & 41,003,731          & \cellcolor[HTML]{1562A9}{\color[HTML]{F1F1F1} 16.5} \\
50 - 64 Years      & 8,820,765      & 63,876,118          & \cellcolor[HTML]{4D99CA}{\color[HTML]{F1F1F1} 13.8} \\
65 - 74 Years      & 3,289,094      & 32,346,340          & \cellcolor[HTML]{B7D4EA}{\color[HTML]{000000} 10.2} \\
75+ Years          & 2,505,606      & 21,790,289          & \cellcolor[HTML]{92C4DE}{\color[HTML]{000000} 11.5} \\ \hline
\end{tabular}

\label{fig:ground_truth}
\end{table}

Based on the infection rate within each age subgroup, a ranking was made ranging from 1, the age group with the lowest infection rate, to 8, the age group with the highest infection rate. A ranking of the sensitivity of the model to each age feature was also determined by first ranking the relative Morris values for each age group for a given delta to obtain a ranking from 1 (lowest) to 8 (highest) value. After ranking each subgroup for all 20 delta values, the average ranking was determined. Both of these rankings, along with the difference between them, are shown in Fig. \ref{table:ranking}.

\begin{table}[!ht]
\centering
\caption{Ranking of the contribution of each age group to the total number of COVID-19 cases based on Infection rate or based on the Morris ranking (Morris), along with the difference between both ranks (Difference).}
\begin{tabular}{|c|c|c|c|}
\hline
\textbf{Age Group} & \textbf{Infection Rank} & \textbf{Morris Rank} & \textbf{Difference} \\ \hline
0 - 4 Years  & \cellcolor[HTML]{08306B}\textcolor{white}{8} & \cellcolor[HTML]{08306B} \textcolor{white}{8}  & \cellcolor[HTML]{FFF5F0} 0                              \\ \hline
5 - 17 Years & \cellcolor[HTML]{2B7BBA} \textcolor{white}{6}  & \cellcolor[HTML]{0B559F} \textcolor{white}{7} & \cellcolor[HTML]{E32F27} \textcolor{white}{1} \\  \hline
18 - 29 Years & \cellcolor[HTML]{F7FBFF} 1                                  & \cellcolor[HTML]{F7FBFF} 1 & \cellcolor[HTML]{FFF5F0} 0                              \\  \hline
30 - 39 Years& \cellcolor[HTML]{DBE9F6} 2                                  & \cellcolor[HTML]{A4CCE3} 3.5 & \cellcolor[HTML]{67000D} \textcolor{white}{1.5}                           \\  \hline
40 - 49 Years& \cellcolor[HTML]{BAD6EB} 3 & \cellcolor[HTML]{DBE9F6} 2 & \cellcolor[HTML]{E32F27} \textcolor{white}{1}                             \\ \hline
 50 - 64 Years& \cellcolor[HTML]{89BEDC} 4 & \cellcolor[HTML]{539ECD} 5                          & \cellcolor[HTML]{E32F27} \textcolor{white}{1}                             \\  \hline
 65 - 74 Years & \cellcolor[HTML]{0B559F} \textcolor{white}{7}                                  & \cellcolor[HTML]{2B7BBA} \textcolor{white}{6}                         & \cellcolor[HTML]{E32F27} \textcolor{white}{1}                            \\  \hline
75+ Years & \cellcolor[HTML]{539ECD} \textcolor{white}{5}                                  & \cellcolor[HTML]{A4CCE3} 3.5  & \cellcolor[HTML]{67000D} \textcolor{white}{1.5}   \\
\hline
\end{tabular}

\label{table:ranking}
\end{table}


Since the average Morris ranking for the age groups 30-39 and 75+ were found to be identical, we chose to break the tie arbitrarily. We noticed that the 5-17 age group ranking of 2 was significantly different from the ground truth ranking of 7. 
However, our results indicate that the Morris ranking closely matches the ground truth ranking for the other seven age groups, as most ranks are identical or have a difference of 1 to 3. This is interpreted as indicating that our predictions of which age groups contribute most to the total COVID-19 case numbers during the time period of the data are approximately equal to the true values verified by CDC and US Census data. 

\section{Related Work}\label{sec:related_works}

Numerous works have been done on COVID-19 forecasting using deep learning and other AI or statistical models \cite{zeroual2020deep, clement2021survey}. Understanding the feature importance and interaction of the input factors is crucial to the adaptation of control measures based on the changing dynamics of the pandemic.

Prior work involved \cite{Malik2022Sensitivity} simulation of the spread of COVID-19 using a fractal-fractional model and used sensitivity analysis to assess the impact of model parameters such as the transmission rate, recovery rate, and vaccination rate on the outcomes of the model. Other work \cite{Ni2022parameter} investigated the impact of key model parameters on the outcomes of a Susceptible-Carrier-Exposed-Infected-Recovered (SCEIR) model predicting COVID-19 infection. Similarly, additional work \cite{monod2021age} analyzed the demographic patterns of COVID-19 cases during the recent resurgence of the pandemic in the United States. The data, consisting of 948 counties, studies the age distribution of cases and their relationship with changes in COVID-19 incidence.


\section{Conclusion and Future Work}\label{sec:conclusion}

Typically, deep learning lacks sufficient interpretability to understand the decisions a model makes to improve predictive performance. However, our work demonstrates the successful application of the Morris method. By analyzing model sensitivity to changes in the age input feature, we explained the impact of COVID-19 on various age group populations in the US between the time period of March 1, 2020 to November 27, 2021. Small perturbations in the static features showed that our models were most sensitive to the 18-29, 30-39, and 40-49 age groups, indicating that these groups contributed most to the total number of COVID-19 cases. These results were then verified by comparison to the ground truth from CDC and Census data. 

After demonstrating the effectiveness of the Morris method applied to deep learning models for understanding the impact of COVID-19 on different age groups, we intend for future experiments to center on exploring the temporal component of these predictions. CDC data show that the ranking of COVID-19 infection rates by age group changes over time. By training new TFT models for a variety of time periods, we can better understand how the impact of COVID-19 varies throughout the pandemic and can further strengthen the reliability of our methods.  

\section{Acknowledgment}
This  work is supported by NSF grant CCF-1918626  Expeditions: Collaborative  Research: Global Pervasive Computational Epidemiology, and NSF Grant 1835631 for CINES: A Scalable Cyberinfrastructure for Sustained Innovation in Network Engineering and Science.

\bibliography{bibliography}
\bibliographystyle{abbrv}

\end{document}